\pdfoutput=1

\documentclass[11pt]{article}

\usepackage{acl2023}

\usepackage{times}
\usepackage{latexsym}
\usepackage{graphicx}
\usepackage{booktabs}
\usepackage{comment}
\usepackage{url}
\usepackage{multirow}
\usepackage{array}
\usepackage{caption}
\usepackage{subcaption}
\usepackage{amsmath}
\usepackage{array}
\usepackage{tabularx}
\usepackage{todonotes}

\usepackage{makecell}
\usepackage{hyperref}
\usepackage{xcolor}
\usepackage{color,soul}
\usepackage{float}
\sethlcolor{yellow}
\usepackage{stfloats,framed}

\usepackage[T1]{fontenc}

\usepackage[utf8]{inputenc}

\usepackage{microtype}

\title{Being Right for {\em Whose} Right Reasons?}

 \author{Terne Sasha Thorn Jakobsen*\textsuperscript{1}\textsuperscript{2}\textsuperscript{3}, Laura Cabello*\textsuperscript{3}, Anders Søgaard\textsuperscript{3} \\
         \textsuperscript{1}Copenhagen Center for Social Data Science \\ \textsuperscript{2}Copenhagen Research Center for Mental Health \\ \textsuperscript{3}University of Copenhagen\\
         \texttt{terne.thorn@sodas.ku.dk}, \texttt{lcp@di.ku.dk}, \texttt{soegaard@di.ku.dk}}

\begin{document}
\maketitle
\def\thefootnote{*}\footnotetext{These authors contributed equally to this work.}\def\thefootnote{\arabic{footnote}}

\begin{abstract}

Explainability methods are used to benchmark the extent to which model predictions align with human rationales
i.e., are `right for the right reasons'. Previous work has failed to 
acknowledge,
however, that what counts as a rationale is sometimes subjective. This paper presents what we think is a first of its kind, a collection of human rationale annotations augmented with the annotators demographic information. 
We cover three datasets 
spanning sentiment analysis and common-sense reasoning, and 
six demographic groups 
(balanced across age and ethnicity). 
Such data enables us to ask both what demographics our predictions align with and
whose reasoning patterns our models' rationales align with. We find systematic inter-group annotator disagreement 
and show how 16 Transformer-based models 
align better with rationales provided by certain demographic groups: 
We find that models are biased towards aligning best with older and/or white annotators. We zoom in on the effects of model size and model distillation, finding -- contrary to our expectations -- 
negative correlations between model size and rationale agreement as well as no evidence that either model size or model distillation improves fairness.

\end{abstract}

\section{Introduction}
Transparency of NLP models is essential for enhancing protection of user rights and improving model performance. A common avenue for providing such insight into the workings of otherwise opaque models come from explainability methods \cite{Paez2019-PEZTPT-3,zednik-boelsen-2022-scientific,Baum2022-BAUFRT-2,https://doi.org/10.1111/phc3.12830,Hacker2022}.  
Explanations for model decisions, also called \textit{rationales}, are extracted to detect when models rely on spurious correlations, i.e., are right for the wrong reasons \cite{mccoy-etal-2019-right}, or to analyze if they exhibit human-like inferential semantics \cite{https://doi.org/10.48550/arxiv.2208.02957,ray-choudhury-etal-2022-machine}. Furthermore, model rationales are used to evaluate how well models' behaviors align with humans, by comparing them to human-annotated rationales, constructed by having annotators mark \textit{evidence} in support of an instance's label \cite{deyoung2019eraser}. Human rationales are, in turn, used in training to improve models by guiding them towards what features they should (or should not) rely on \cite{Mathew2021,rajani2019explain}.

\begin{figure}
    \centering
    \includegraphics[width=\linewidth]{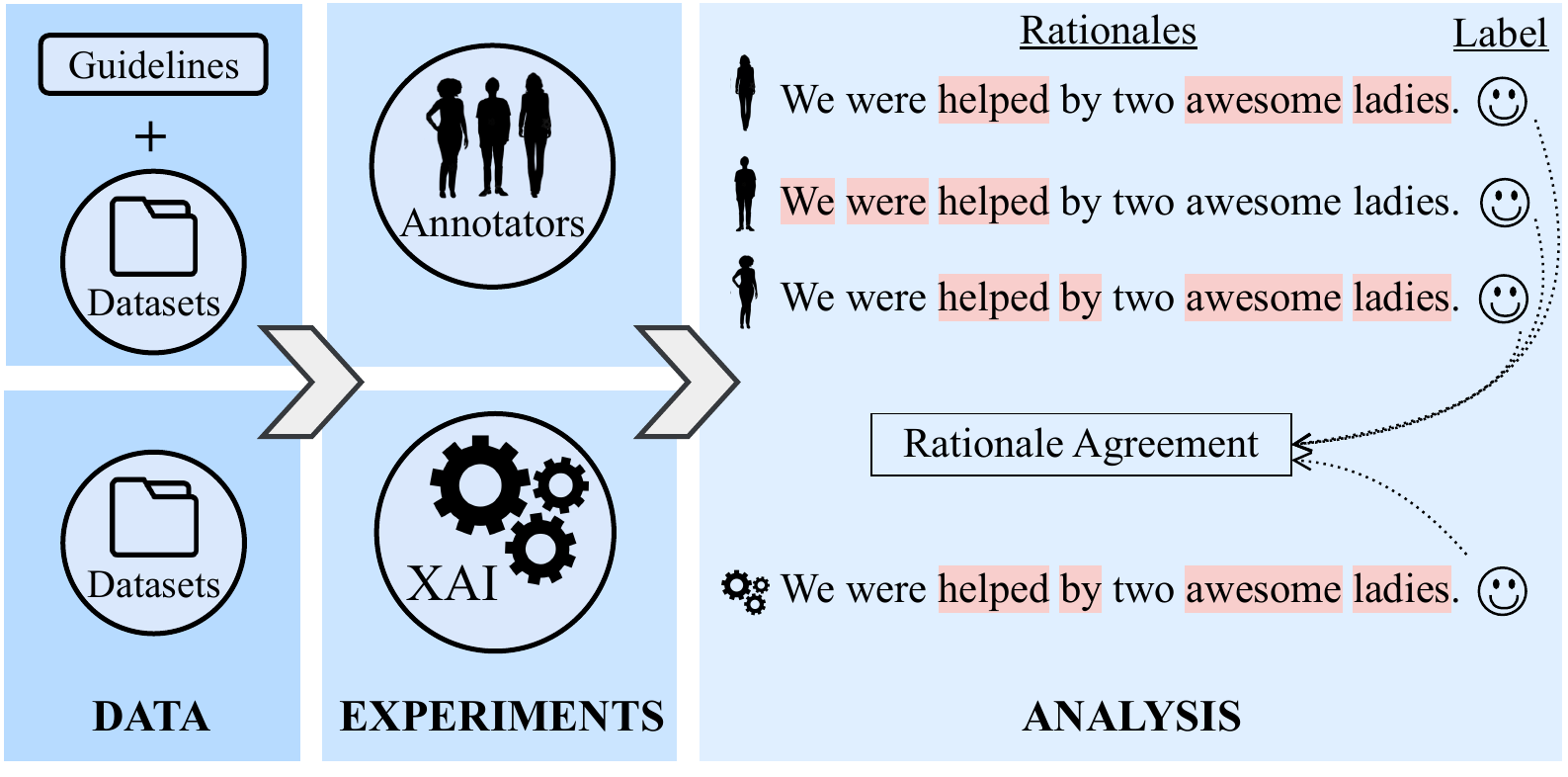}
    \caption{Experimental setup for a sentiment analysis task. For a given instance, annotators are asked to choose a label and mark supporting evidence for their choice. For instances with full label agreement, we compare alignment of rationales (group-group alignment). We do the same to measure group-model alignment through attention- and gradient-based explainability methods.}
    \label{fig:overview}
\end{figure}

While genuine disagreement in labels is by now a well-studied phenomenon \cite{beigman-klebanov-beigman-2009-squibs,plank-etal-2014-learning,Plank2022TheO}, little attention has been paid to disagreement in rationales.
Since there is evidence that human rationales in ordinary decision-making differ across demographics \cite{Stanovich2000}, we cannot, it seems, blindly assume that what counts as a rationale for one group of people, e.g,. young men, also counts as a rationale for another group of people, e.g., elderly women. This dimension has not been explored in fairness research either. Could it be that some models that exhibit performance parity, condition on factors that align with the rationales of some groups, but not others?

\paragraph{Contributions} 
We present a collection of three existing datasets with demographics-augmented annotations to enable profiling of models, i.e., quantifying their alignment\footnote{We use the terms `agreement' and `alignment' interchangeably.} with rationales provided by different socio-demographic groups. Such profiling enables us to ask {\em whose}~right reasons models are being right for. Our annotations span two NLP tasks, namely \textit{sentiment classification} and \textit{common-sense reasoning}, across three datasets and six demographic groups, defined by age \{Young, Old\} and ethnicity \{Black/African American, White/Caucasian, Latino/Hispanic\}. We investigate label and rationale agreement across groups and evaluate to what extent groups' rationales align with 16 Transformer-based models' rationales, which are computed through attention- and gradient-based methods.
We observe that models generally align best with older and/or white annotators. While larger models have slightly better prediction performance, model size does not correlate positively with neither rationale alignment nor fairness.
Our work constitutes multi-dimensional research in off-the-beaten-track regions of the NLP research manifold \cite{ruder-etal-2022-square}. We make the annotations publicly available.
\footnote{\href{https://github.com/terne/Being_Right_for_Whose_Right_Reasons}{\nolinkurl{github.com/terne/Being_Right_for_Whose_Right_Reasons}}}
\footnote{\href{https://huggingface.co/datasets/coastalcph/fair-rationales}{\nolinkurl{huggingface.co/datasets/coastalcph/fair-rationales}}}

\section{Fairness and Rationales} 

Fairness generally concerns the distribution of resources, often across society as a whole. In NLP, the main resource is system performance. Others include computational resources, processing speed and user friendliness, but {\em performance is king}. AI fairness is an attempt to regulate the distribution of performance across subgroups, where these are defined by the product of legally protected attributes \cite{pmlr-v97-williamson19a}.

NLP researchers have uniformly adopted American philosopher John Rawls' definition of fairness \cite{larson-2017-gender,NEURIPS2020_92650b2e,ethayarajh-jurafsky-2020-utility,NEURIPS2021_908075ea,chalkidis-etal-2022-fairlex}, defining fairness as performance parity, except where it worsens the conditions of the least advantaged. Several dozen metrics have been proposed, based on Rawls' definition \citep{castelnovo2022}, some of which are argued to be inconsistent or based on mutually exclusive normative values 
\cite{10.1145/3433949,castelnovo2022}. \citet{10.1145/3194770.3194776} grouped these metrics into metrics based only on predicted outcome, e.g., statistical parity, and metrics based on both predicted and actual outcome, e.g., performance parity and accuracy equality. \citet{CorbettDavies2018TheMA} argue that metrics such as predictive parity and accuracy equality do not track fairness in case of infra-marginality, i.e., when the error distributions of two subgroups are different. For a better understanding of the consequences of infra-marginality 
we refer to \citet{DBLP:journals/corr/abs-1909-00982} and \citet{sharma2020inframarginality}. Generally, there is some consensus that fairness in NLP is often best evaluated in terms of performance parity using standard performance metrics \cite{pmlr-v97-williamson19a,koh2020wilds,chalkidis-etal-2022-fairlex, ruder-etal-2022-square}.
We do the same and evaluate fairness in group-model rationale agreement quantifying performance differences (understanding performance as degree of rationale agreement) across end user demographics. In doing so, we are embodying group fairness values: that individuals should be treated equally regardless of their protected attributes, i.e., group belonging.

Fairness and explainability are often intertwined in the literature due to the assumption that transparency, through explainability methods, makes it possible to identify which models are right for the right reasons or, on the contrary, right by relying on spurious, potentially harmful, patterns \cite{DBLP:journals/ai/LangerOSHKSSB21,balkir-etal-2022-challenges}. This study tightens the connection between fairness and explainability, investigating whether model rationales align better with those of some groups rather than others. If so, this would indicate that models can be more robust for some groups rather than others, even in the face of performance parity on dedicated evaluation data. That is: We ask whether models are equally right for the right reasons (with the promise of generalization) across demographic groups.

\begin{figure*}[t]
    \centering
    \includegraphics[width=\linewidth]{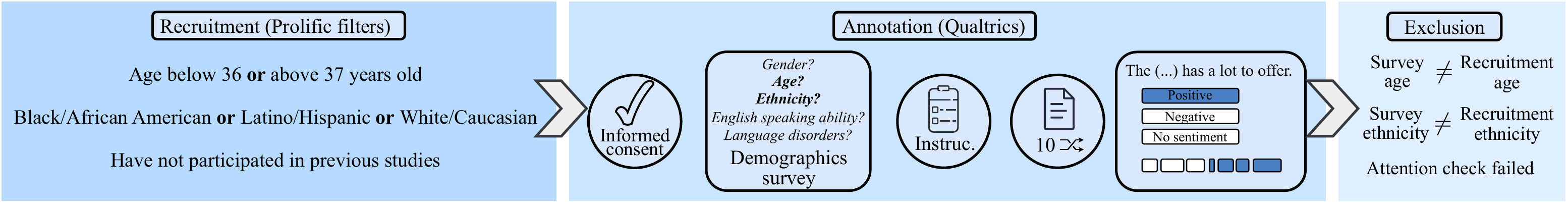}
    \caption{Overview of the annotation collection process from annotator recruitment criteria, to the annotation itself, and finally annotator exclusion criteria. Separately for each dataset, annotators are recruited via Prolific using specific filters for age, ethnicity and participation status. Recruits are directed to a Qualtrics survey containing, in consecutive order, a consent form, a short demographics survey, instructions for the annotation task and then approx. 10 randomly selected instances of which annotators provide both labels and rationales for. After annotation, some annotators' responses are excluded from our analysis due to certain mismatches in responses. The annotation process is detailed further in section \ref{sec:annotation_process} and we show the instructions and task examples in appendix A.}
    \label{fig:annotation_overview}
\end{figure*}

\section{Data}

We augment a subset of data from three publicly available datasets spanning two tasks: DynaSent \cite{potts-etal-2020-DynaSent} and SST \cite{socher2013recursive}\footnote{We work with its binary version, SST-2.}, 
for sentiment classification and CoS-E \cite{talmor-etal-2019-commonsenseqa,rajani2019explain} for common-sense reasoning.\footnote{We use the simplified version of CoS-E released by \citet{deyoung2019eraser}.}
For each dataset, we crowd-source annotations for a subset of the data. 
We instruct annotators to select a label and provide their rationale for their choice by highlighting supporting words in the given sentence or question. 
Table~\ref{tab:datasum} shows statistics of the annotations collected. Annotation guidelines are explained in \S~\ref{sec:annotation_process} (and included in full in Appendix~\ref{app:annotation-guidelines}) and recruitment procedures are explained in \S~\ref{sec:recruitment}.

\begin{table}[ht!]
    \centering
    \resizebox{\columnwidth}{!}{
    \begin{tabular}{lccc}
    \toprule
          
          & \multicolumn{2}{c}{Annotators} & Annotations \\
          \cmidrule(r){2-3}\cmidrule(r){4-4}
          & $\times$Group & Total & Total \\
          \midrule
          {\sc DynaSent}& 48 & 288 & 2,880\\
          {\sc SST-2}& 26 & 156 & 1,578\\
          {\sc CoS-E}& 50 & 300 & 3,000\\
          \midrule
          \textbf{{\sc Total}} & \textbf{124} & \textbf{744} &\textbf{7,458}\\
          {\sc Before excl.*} & - & 929 & 9,310\\
         \bottomrule
    \end{tabular}}
    \caption{Summary of the annotated data, showing, for each dataset, the amount of annotators within the six demographic groups, the total amount of annotators and the amount of annotations after workers have annotated approx. 10 instances each. Reported numbers are after exclusions as described in \S~ \ref{sec:recruitment}. *We publicly share all annotated data which includes annotators that were excluded from our analyses.
    }
    \label{tab:datasum}
\end{table}

\subsection{Annotation Process}
\label{sec:annotation_process}

We summarize the process of collecting annotations in Figure \ref{fig:annotation_overview}, where we depict a three-step process: recruitment, annotation and exclusion. In this section, we start by describing the second step -- annotation -- and explain \textit{what} is annotated and \textit{how} it is annotated. We describe our recruitment and exclusion criteria in the following section, \ref{sec:recruitment}. 

Annotators are directed to a Qualtrics\footnote{\url{https://www.qualtrics.com}} survey and presented with \emph{i)} a consent form, \emph{ii)} a short survey on demographics, \emph{iii)} instructions for their annotation task and lastly, \emph{iv)} a randomly selected set of $n\approx10$ instances to annotate, out of a subset of size $N$. As a result of this procedure, each group, for each dataset, is represented by approximately $N/n$ annotators. 
Data points are annotated for both classification labels and extractive rationales, i.e., input words that motivate the classification. 

Existing rationale datasets are typically constructed by giving annotators `gold standard' labels, and having them provide rationales for these labels. Instead, we let annotators provide rationales for labels they choose themselves. This lets them engage in the decision process, but it also acknowledges that annotators with different backgrounds may disagree on classification decisions.  
Explaining other people's choices is error-prone \cite{BaraszKate2022CpMs}, and we do not want to bias the rationale annotations by providing labels that align better with the intuitions of some demographics than with those of others.  
For the sentiment analysis datasets, we discard neutral instances because rationale annotation for neutral instances is ill-defined.  
Yet, we still allow annotators to evaluate a sentence as neutral, since we do not want to force our annotators to provide rationales for positive and negative sentiment that they do not see.

\paragraph{DynaSent} We re-annotate $N=480$ instances six times (for six demographic groups), comprising 240 instances labeled as positive, and 240 instances labeled as negative in the DynaSent Round 2 test set (see \citet{potts-etal-2020-DynaSent}). This amounts to 2,880 annotations, in total. Our sentiment \textit{label} annotation follows the instructions of \citet{potts-etal-2020-DynaSent}. To annotate \textit{rationales}, we formulate the task as marking ``supporting evidence'' for the label, following how the task is defined by \citet{deyoung2019eraser}. Specifically, we ask annotators to mark all the words, in the sentence, they think shows evidence for their chosen label.

\paragraph{SST-2} We re-annotate $N=263$ instances six times (for six demographic groups), which are all the positive and negative instances from the Zuco dataset of \citet{Hollenstein2018ZuCoAS}\footnote{The Zuco data contains eye-tracking data for 400 instances from SST. By annotating some of these with rationales, we add an extra layer of information for future research. Note that there is a typo in \cite{Hollenstein2018ZuCoAS}.  
There is 263 positive and negative instances (not 277).}, comprising a mixture of train, validation and test set instances from SST-2, which we remove from the original data before training the models.
Instructions for sentiment annotations build on 
the instructions by \citeauthor{potts-etal-2020-DynaSent}, combined with a few examples from \citet{zaidan-etal-2007-using}. The instructions for annotating rationales are the same as for DynaSent.

\paragraph{CoS-E} We re-annotate $N=500$ instances from the test set six times (for six demographic groups) and ask annotators to firstly select the answer to the question that they find most correct and sensible, and then mark words that justifies that answer. Following \citet{chenghan2022}, we specify the rationale task with a wording that should guide annotators to make short, precise rationale annotations: \begin{quote}`For each word in the question, if you think that removing it will decrease your confidence toward your chosen label, please mark it.'\end{quote}

\subsection{Annotator Population}\label{sec:recruitment} 

We recruited annotators via Prolific based on two main criteria, age and ethnicity, previously identified as related to unfair performance differences of NLP systems \cite{hovy-sogaard-2015-tagging,jorgensen-etal-2016-learning,sap-etal-2019-risk,zhang-etal-2021-sociolectal}. 

\paragraph{Recruitment} In our study, there is a trade-off between collecting annotations for a diverse set of data instances (number of tasks and sentences) and for a diverse set of annotators (balanced by demographic attributes), while keeping the study affordable and payment fair. Hence, when we want to study differences between individuals with different ethnic backgrounds, we can only study a subset of possible ethnic identities (of which there are many categories and diverging definitions). We balanced the number of annotators across \textit{three} ethnic groups --- Black/African American (B), Latino/Hispanic (L) and White/Caucasian (W) --- and \textit{two} age groups ---below 36 (young, Y) and above 37 (old, O), excluding both --- whose cross-product results in six sub-groups: \{BO, BY, LO, LY, WO, WY\}. We leave a two-year gap between the age groups in order to not compare individuals with very similar ages. Furthermore, the age thresholds are inspired by related studies of age differences in NLP-tasks and common practices in distinguishing groups with an age gap \cite{johannsen-etal-2015-cross, hovy-sogaard-2015-tagging} and around the middle ages \cite{zhang-etal-2021-sociolectal}. Our threshold also serves to guarantee sufficient proportions of available crowdworkers in each group. Our ethnicity definition follows that of Prolific, which features in a question workers have previously responded to and hence are recruited by, defining ethnicity as: 
\begin{quote}`[a] feeling of belonging and attachment to a distinct group of a larger population that shares their ancestry, colour, language or religion'\end{quote}

While we do not require all annotators to be fluent in English, we instead ask about their English-speaking abilities in the demographics survey and find that 75\% of the participants speak English “very well” and only 1\% “not well”, and the remaining ``well''.

\paragraph{Exclusions} Annotators who participated in annotating one task were excluded from participating in others. \textit{After} annotation, we manually check whether a participant's answers to our short demographics survey correspond to their recruitment criteria. We found many discrepancies between recruitment ethnicity and reported ethnicity, especially for Latino/Hispanic individuals, who often report to identify as White/Caucasian. This highlights the difficulty of studying ethnicities as distinct, separate groups, as it is common to identify with more than one ethnicity\footnote{General Social Survey as well as US Census allow respondents to report multiple ethnicities for this reason. See, e.g., a GSS 2001 report commenting on multi-ethnicity: \url{shorturl.at/BCP49}.}. Hence, the mismatches are not necessarily errors. For our experiments, we decided to exclude participants with such mismatches and recruit new participants to replace their responses  
(see Appendix~\ref{app:annotations-overview} for further details). 
A smaller amount of participants were excluded due to mismatch in reported age or due to failing a simple attention check. We release annotations both with and without the instances excluded from our analyses.
The final data after pre-processing consist of one annotation per instance for each of the six groups, i.e., six annotations per instance in total. Annotators annotated (approximately) 10 instances each. \textit{All} participants were paid equally.

\section{Experiments}
\label{sec:experiments}

We first conduct an analysis of \textit{group-group} label agreement (i.e., comparing human annotator groups with each other, measuring human agreement on the sentiment and answer labels) and rationale agreement (measuring human agreement on rationale annotations) to characterize inter-group differences. 
We then move to \textit{group-model} agreement (comparing the labels and rationales of our annotator groups to model predictions and model rationales) and ask: 
Do models' explanations align better 
with certain demographic groups compared to others? In our analysis, we further focus on how rationale agreement and fairness behave depending on model size and model distillation. 

We probe 16 Transformer-based models\footnote{All pretrained models can be downloaded at \href{https://huggingface.co/models}{\nolinkurl{huggingface.co/models}}.}. To ease readability, we will use abbreviations following their original naming when depicting models' performance\footnote{\{{\tt abv2}: albert-base-v2, {\tt alv2}: albert-large-v2, {\tt mlm-l6}: MiniLM-L6-H384-uncased, {\tt mlm-l12}: MiniLM-L12-H384-uncased, {\tt axlv2}: albert-xlarge-v2, {\tt dbu}: distilbert-base-uncased, {\tt dr}: distilroberta-base, {\tt bbu}: bert-base-uncased, {\tt rb}: roberta-base, {\tt mrb}: muppet-roberta-base, {\tt dv3b}: deberta-v3-base, {\tt axxlv2}: albert-xxlarge-v2, {\tt blu}: bert-large-uncased, {\tt rl}: roberta-large, {\tt mrl}: muppet-roberta-large, {\tt dv3l}: microsoft/deberta-v3-large\}}.

\begin{figure}[ht!]
    \centering
    \includegraphics[width=\linewidth]{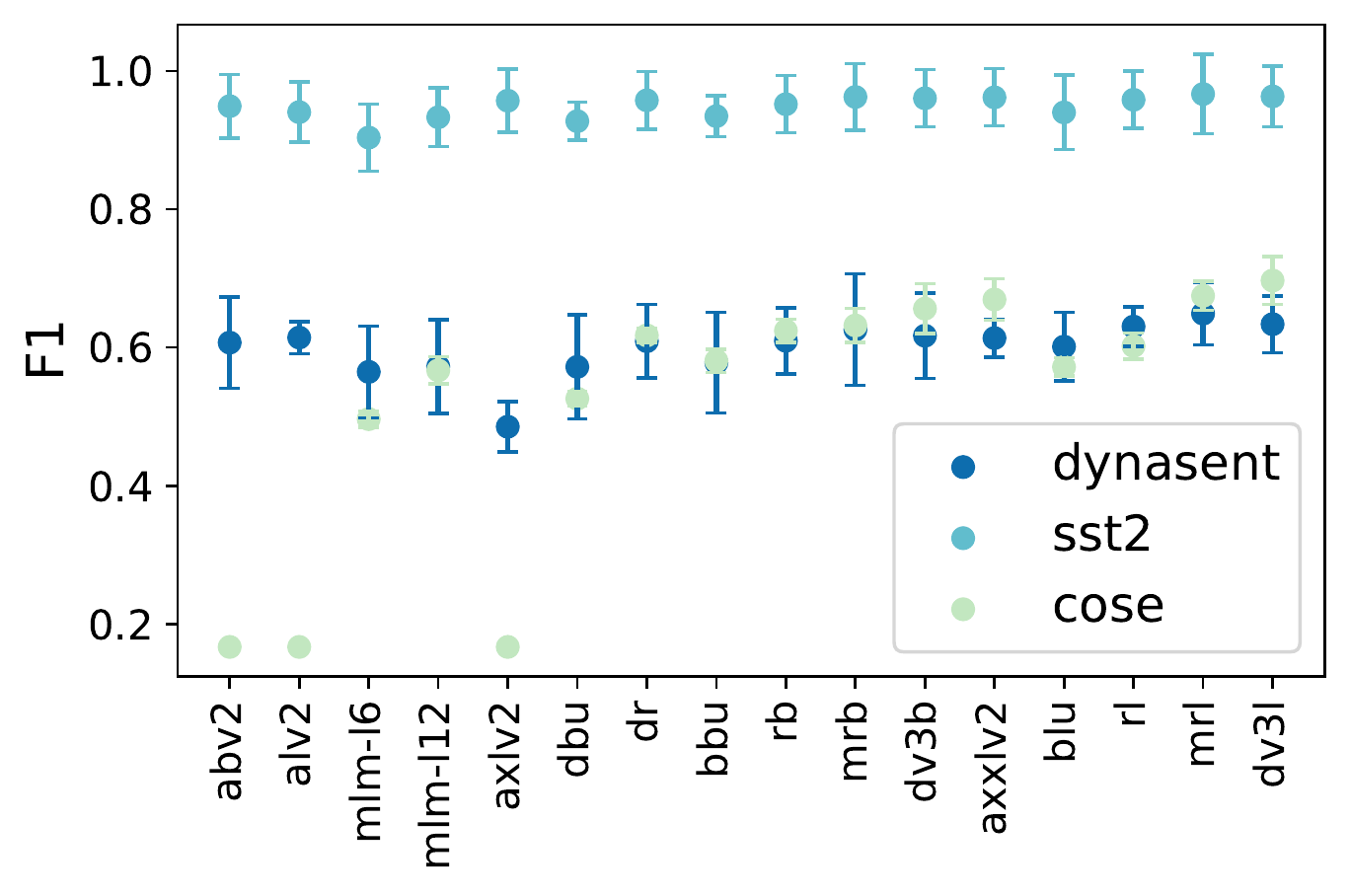}
    \caption{Group-model label agreement over our annotated data, measured by F1-score. Error bars show variance between the best and worst performing groups. Models are ordered by size from smallest to largest from left to right.}
    \label{fig:f1-scsores}
\end{figure}

\noindent We fine-tune the models individually on each dataset (see Figure~\ref{fig:f1-scsores}).
SST-2 and CoS-E simplified\footnote{CoS-E simplified represents each of the original questions into five question-answer pairs, one per potential answer, and label them as True (the right question-answer pair) or False.\label{footcose}} are modeled as binary classification tasks; DynaSent is modeled as a ternary (positive/negative/neutral) sentiment analysis task. We exclude all annotated instances from the training splits; for CoS-E, we downsample the negative examples to balance both classes in the training split. After fine-tuning for 3 epochs, we select the checkpoint with the highest validation accuracy to run on our test (annotated) splits and apply two explainability methods 
to obtain input-based explanations, i.e., rationales, for the predictions made.

We measure label agreement with appropriate variants of F$_1$ (SST-2 binary-F$_1$; DynaSent macro-F$_1$; CoS-E mean of binary-F$_1$ towards the negative and the positive class). CoS-E simplified represents a slightly different task (see footnote~\ref{footcose}) from what the annotators were presented to solve (a multi-class question-answering task). To correctly measure label agreement, we evaluate whether a model predicts `True' for the question-answer pair with the answer selected by the annotator. Therefore, to avoid misleading F$_1$ scores if, for example, a model predominantly predict True, we report the mean of the F$_1$ towards each class. We explain below how we measure rationale agreement.

\paragraph{Explainability methods}

We analyze models' predictions through two families of post-hoc, attribution-based\footnote{The methods are applied at inference time 
and provide explanations \textit{locally}, i.e., for each individual instance, 
indicating the relative importance of each input token through a score distribution.
} explainability methods: Attention Rollout (AR) \cite{abnar-zuidema-2020-quantifying} 
and Layer-wise Relevance Propagation (LRP) \cite{bach-lrp}, a gradient-based method. \citet{xai-lrp} compare these methods, showing how their predicted rationales are frequently uncorrelated.  
Both AR and LRP thus provide token level rationales for a given input, but while AR approximates the relative importance of input tokens by accumulating attention, LRP does so by backpropagating `relevance' from the output layer to the input, 
leading to sparser attribution scores.
We rely on the rules proposed in \citet{xai-lrp}, an extension of the original LRP method \cite{bach-lrp,arras-etal-2017-explaining} for Transformers, aiming to uphold the conservation property of LRP in Transformers as well. This extension relies on an ``implementation trick'', whereby the magnitude of any output remains intact during backpropagation of the gradients of the model.

\paragraph{Comparing rationales}

Attention-based and gradient-based methods do not provide categorical relevance of the input tokens, but a vector $S_i$ with continuous values for each input sentence $i$. We translate $S_i$ into a binary vector $S_i^b$ following the procedure from \citet{Wang2022AFI} for each group. We define the top-$k^{gd}$ tokens as rationales, where $k^{gd}$ is the product of the current sentence length (tokens) and the average rationale length ratio (RLR) of a group $g$ within a dataset $d$. 
On average, RLR for SST-2 are shorter (29.6\%) compared to DynaSent (31.9\%) and CoS-E (33.0\%) (see Appendix~\ref{app:annotations-overview} for specific values). 
Models' outputs are also preprocessed to normalize different tokenizations and to match the input format given to annotators.

After aligning explanations from models and annotators in the same space, we can compare them. We employ
two metrics specifically designed to evaluate discrete rationales: token-level $\mathrm{F_1}$ ($\mathrm{\text{token-}F_1}$) (Equation~\ref{eq:tokenf1}) \cite{deyoung2019eraser, Wang2022AFI}, and Intersection-Over-Union $\mathrm{F_1}$ ($\mathrm{\text{IOU-}F_1}$) (Equation~\ref{eq:iouf1}) as presented in \cite{deyoung2019eraser}. These metrics are flexible enough to overcome the strictness of exact matching.\footnote{Formally,

\begin{equation}\label{eq:tokenf1}
\text{token-}F_1 = \frac{1}{N}\sum_{i=1}^N2\times \frac{P_i\times R_i}{P_i+R_i}
\end{equation}

\noindent where $P_i$ and $R_i$ are the precision and recall for the $i^{th}$ instance,
computed by considering the overlapped tokens between models' and annotators' rationales. To measure Intersection-Over-Union, we define the categorical vector given by the annotators for each sample as $A_i$. Thereby,

\begin{equation}\label{eq:iou}
\text{IOU}_i = \frac{|S_i^b \cap A_i|}{|S_i^b \cup A_i|}
\end{equation}

\noindent and 

\begin{equation}\label{eq:iouf1}
\text{IOU-}F_1 = \frac{1}{N}\sum_{i=1}^N \left\{
    \begin{array}{ll}
      1 & \mbox{if $\text{IOU}_i\geq 0.5$} \\
      0 & \mbox{otherwise}.
    \end{array}
  \right. 
\end{equation}

These metrics account for \emph{plausibility} \cite{deyoung2019eraser} of the models' rationales, i.e., the degree to which they are agreeable to humans, as well as the extent to which models are `right for the right reasons' \cite{mccoy-etal-2019-right}. Since we are interested in comparing rationale alignment between groups and between groups and models, 
measuring plausability is our go-to. Other research \cite{jacovi-goldberg-2020-towards, SETZU2021103457} 
focus on properties like \emph{faithfulness}, which reflect a model's true decision process, i.e., whether the provided rationale influenced the corresponding decision, generally measured through perturbation experiments.}

\begin{figure*}
    \centering
    \resizebox{\textwidth}{!}{
    \includegraphics[width=0.3\linewidth]{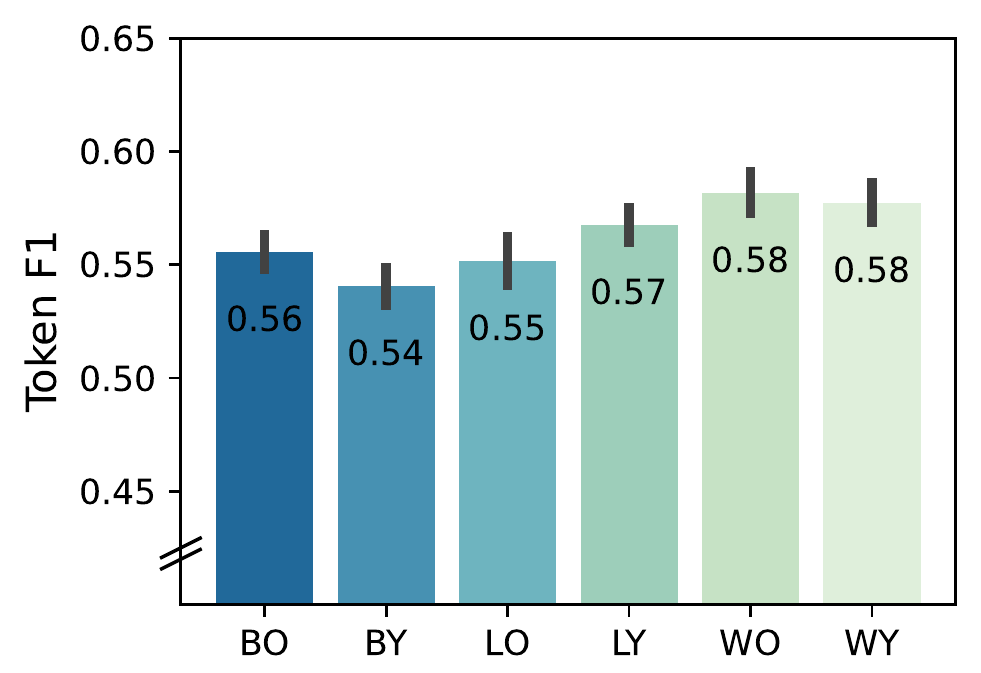}
    \hspace{0.5cm}
    \includegraphics[width=0.65\linewidth]{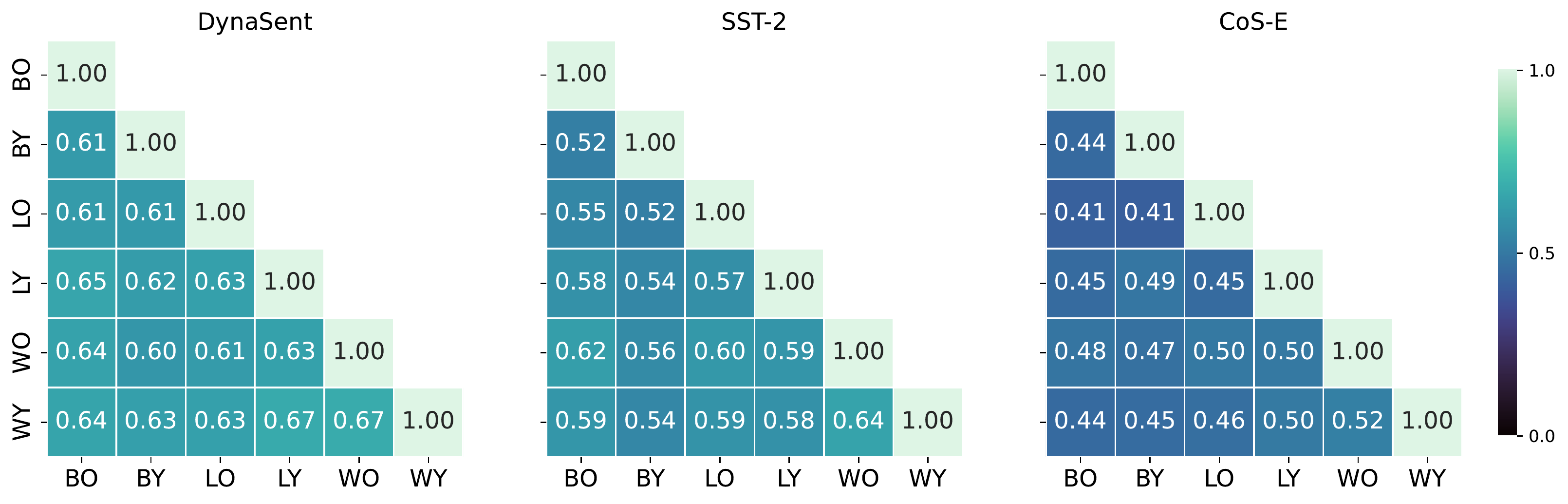}
    }
    \caption{Group-group \textbf{rationale agreement} for instances 
    with full label agreement.   
    Agreement is measured by token-level binary $\mathrm{F_1}$. On the left side, average and std (error bar) $\mathrm{\text{token-}F_1}$ for 20 random combinations of paired group rationales over all datasets. On the right, each group-group agreement for each dataset.
    We observe lower agreement for BY except in CoS-E. WO tends to agree more with other groups, especially in CoS-E.}
    \label{fig:groupgroup_agreement}
\end{figure*}

\section{Results and Discussion}\label{sec:results}

Figure~\ref{fig:f1-scsores} shows group-model label agreement over our annotated data.\footnote{See Figure~\ref{fig:model_group_f1} in Appendix~\ref{app:supplementary-figures} for a detailed representation of group-model label agreement.} Error bars show the variability between best and worst performing groups. CoS-E exhibits the lowest variability, indicating less variability in label agreement between groups.

When annotators disagree on the label of an instance, it is to be expected that their rationales will subsequently be different. Therefore, to compare group-group (\S~\ref{sec:analysis-groupgroup}) and group-model (\S~\ref{sec:analysis-modelgroup}) rationales more fairly, we focus on the subset of instances where all groups are in agreement about the label, i.e., instances with full label agreement. This amounts to 209, 152 and 161 instances for DynaSent, SST-2 and CoS-E, respectively.

\subsection{Analysis of Group-Group Agreement}\label{sec:analysis-groupgroup}

We first want to quantify how different the rationales of one group are to those of others, and more generally to a random population. 
We compare each groups' set of rationales to a random paired set of rationales, where the rationale of each instance is randomly picked from one of the five other groups. Figure \ref{fig:groupgroup_agreement} shows the overall agreement score, average $\mathrm{\text{token-}F_1}$ across datasets, and its standard deviation from 20 random seeds, i.e., 20 random combinations of paired rationales. 
We observe that rationales of White annotators (WO, WY) are on average more similar to others while the average difference with the rationales of minority groups like, for example, Black Young (BY), is greater.

We then compute the level of rationale agreement ($\mathrm{\text{token-}F_1}$) between all groups (heatmaps on Figure~\ref{fig:groupgroup_agreement}) and observe that, in general, differences in group-group rationale agreement 
are consistent across datasets (tasks): Black Youngs (BY) have lower alignment with others, especially in sentiment analysis tasks. 
While the definition of rationales for DynaSent seems to be easier (higher values of agreement), it seems to be harder (lower values of agreement) for CoS-E, even when the label is agreed upon.
We hypothesize this is due to the complexity of the CoS-E task itself, which also leads to more lengthy rationales, as reflected by the average RLR reported on \S~\ref{sec:experiments},
probably in the absence of a clear motivation for the selected answer. 

The definition of what is \emph{common-sense} varies across cultures and it is related to a person's background \cite{hershcovich-etal-2022-challenges}, which makes CoS-E a highly subjective task\footnote{This is specially notorious on the query type \emph{people}.}. Take for example the question `Where would you find people standing in a line outside?' with these potential answers: `bus depot', `end of line', `opera', `neighbor's house' and `meeting'. Even if there is agreement on the \emph{correct} choice as `bus depot', the rationale behind it could easily differ amongst people, i.e., it could be due to `people standing', or the fact that they are standing in `a line outside', or all together.

\begin{figure}[ht!]
    \centering
    \includegraphics[width=0.9\linewidth]{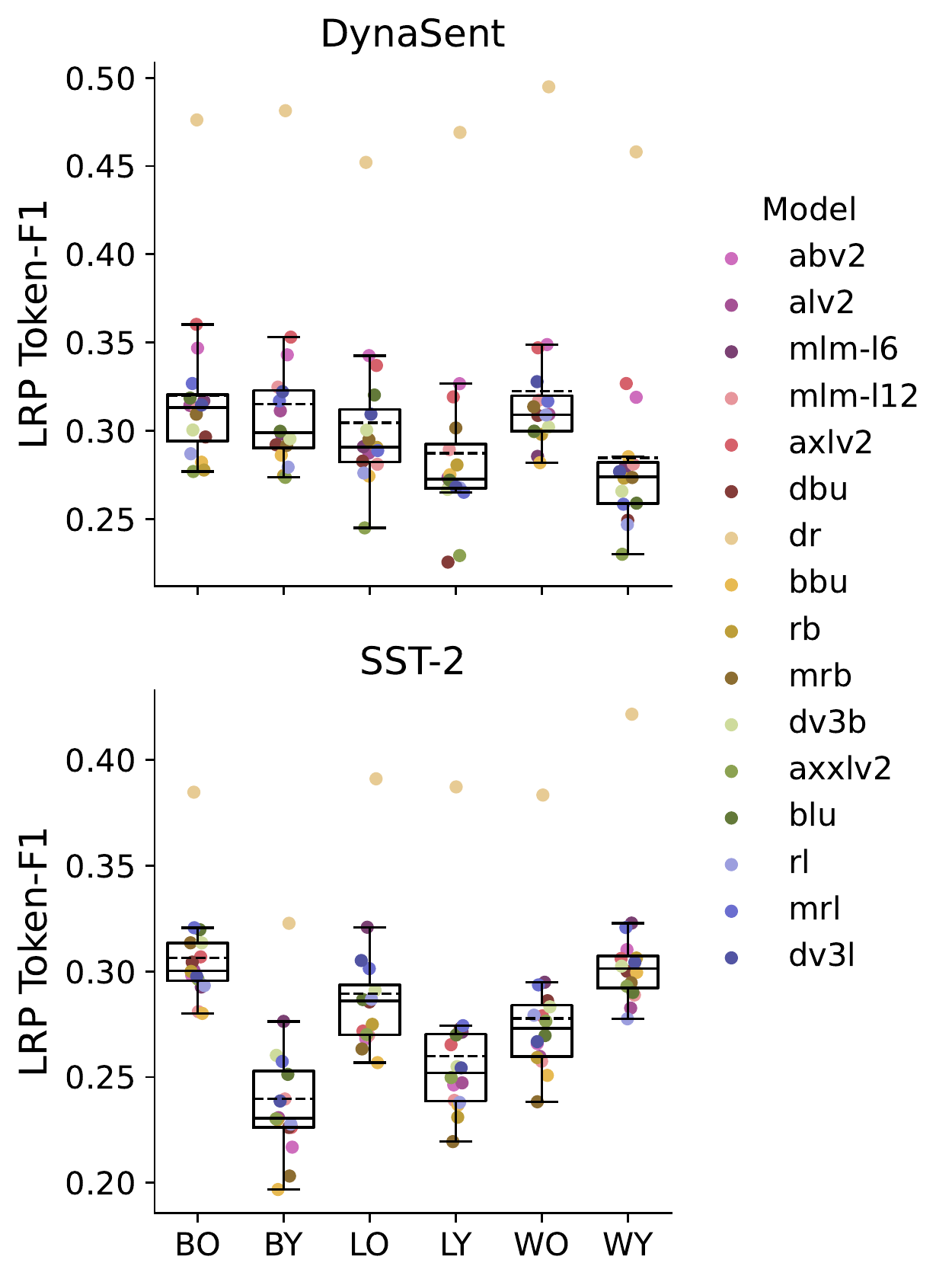}
    \setlength{\belowcaptionskip}{-0.5cm}
    \caption{Box-plots of group-model rationale alignment for the two sentiment datasets measured with $\mathrm{\text{token-}F_1}$. Model rationales are extracted with LRP. Each dot represents a model's $\mathrm{\text{token-}F_1}$ score for the respective group. We see that for each ethnic group, model rationales align better with rationales of older annotators, except for White Young (WY) annotators of SST-2. DistilRoBERTa ({\tt dr}) is an outlier, consistently showing the best scores in both datasets across groups.}
    \label{fig:tokenf1_boxplot}
\end{figure}

\subsection{Analysis of Group-Model Agreement}\label{sec:analysis-modelgroup}
Now that we have analyzed group-group agreement, we measure the alignment between groups' rationales and models' rationales. We analyze predictions from 16 Transformer-based models and employ AR and LRP to extract model rationales. Methods for comparing rationales and measuring group-model agreement are explained in Section~\ref{sec:experiments}.

\begin{figure*}
    \centering
    \includegraphics[width=\textwidth]{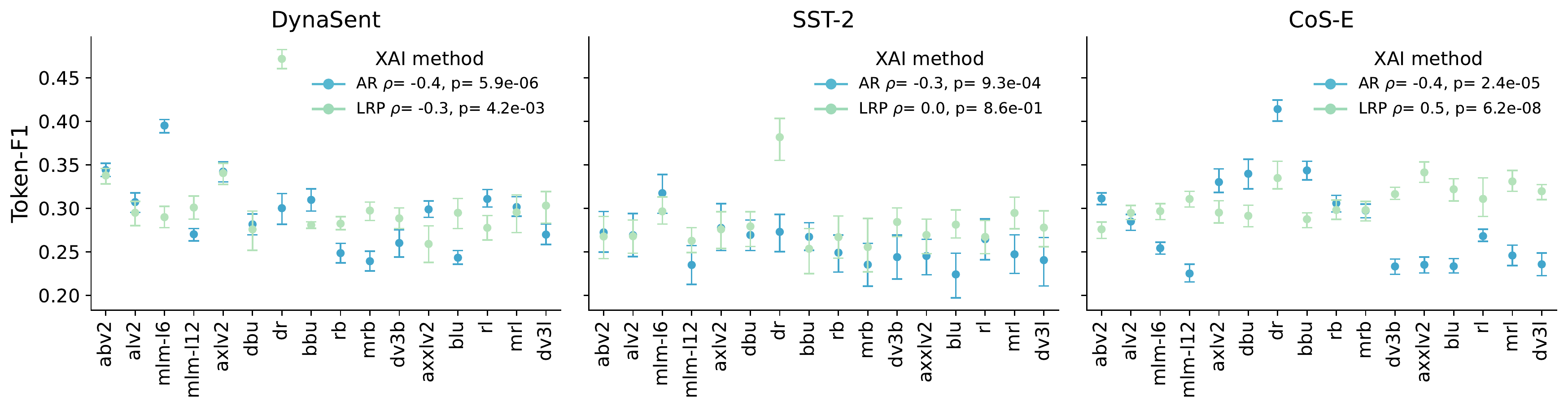}
    \caption{Group-model rationale alignment ($\mathrm{\text{token-}F_1}$). Error bars show the distance between the groups with the highest and lowest scores. On the X-axis, models are ordered from smallest to largest. 
    We show Spearman correlation coefficients, $\rho$, between $\mathrm{\text{token-}F_1}$ scores (the concatenation of all groups' scores) and model sizes (in Million parameters), finding $\mathrm{\text{token-}F_1}$ to be negatively correlated with model size in most cases.} 
    \label{fig:modelsize_tokenf1}
\end{figure*}

\paragraph{Socio-demographic fairness}
Figure \ref{fig:tokenf1_boxplot} shows a systematic pattern of model rationales aligning better with the rationales of older annotators in each ethnic group (BO, LO, WO) on the sentiment datasets. The only exception is White Young (WY) annotators in SST-2, whose median $\mathrm{\text{token-}F_1}$ is higher than their older counterpart. 
We argue this is due, in part, to the data source of the tasks themselves. While DynaSent constitutes an ensemble of diverse customer reviews, SST is based on movie review excerpts from Rotten Tomatoes with a more informal language, popular amongst younger users. Findings from \citet{johannsen-etal-2015-cross} and \citet{hovy-sogaard-2015-tagging} indicate that there exist grammatical differences between age groups. \citet{johannsen-etal-2015-cross} further showed several age and gender-specific syntactic patterns that hold even across languages. This would explain not only the noticeable group-group differences when marking supporting evidence (lexical structures) for their answers, but also the agreement disparity reflected by models fine-tuned on potentially age-biased data. 

Results are consistent with previous findings of \citet{zhang-etal-2021-sociolectal}, who show a variety of language models aligning better with older, white annotators, and worse with minority groups, in word prediction tasks. We observe that group-model rationale agreement does not correlate with group-model class agreement, i.e., when a model performs well for a particular group, it does not necessarily entail that its rationales, or learned patterns, align. 
Group-model rationale agreement evaluated with Attention Rollout and CoS-E are shown in Figure~\ref{fig:complete-group-model-alignment} in Appendix~\ref{app:supplementary-figures}, along with results using the complementary metric ($\mathrm{\text{IOU-}F_1}$). The patterns derived from them are in line with those in Figure~\ref{fig:tokenf1_boxplot}: AR shows similar behaviours as LRP, but leads to larger variation between models. However CoS-E, which, as explained, is a very different task, does not seem to exhibit big group differences. This is also noticeable in Figure~\ref{fig:modelsize_tokenf1}, where error bars show the distance between groups with the highest and lowest level of agreement in every model.

\paragraph{The role of model size}
In general, larger language models seem to perform better on NLP tasks. In our setting, Figure~\ref{fig:f1-scsores} shows a positive trend with model size: larger models achieve, in general, higher performance. Could it be the case that larger language models 
also show higher rationale agreement? And, are they consequently more fair? We evaluate fairness in terms of performance parity: min-max difference between the group with the lowest and highest $\mathrm{\text{token-}F_1}$ (per model). 
Relying on min-max difference captures the widely shared intuition that fairness is always in the service of the worst off group \cite{rawls_theory_1971}.

Contrary to our expectations, Figure~\ref{fig:modelsize_tokenf1} shows how $\mathrm{\text{token-}F_1}$ scores actually \textit{decrease} with model size -- with CoS-E model rationales from LRP being the only exception to the trend. We report Spearman correlation values for each dataset and explainability method: The negative correlation between $\mathrm{\text{token-}F_1}$ and model size is significant in all three datasets with AR, but only in DynaSent with LRP. The positive correlation in CoS-E with LRP rationales is also significant. 

When we zoom in on the min-max $\mathrm{\text{Token-}F_1}$ gaps (error bars on Figure \ref{fig:modelsize_tokenf1})\footnote{See Figure~\ref{fig:minmax} in Appendix~\ref{app:rationale-agreement} for a plot of the gaps themselves.}, we find that performance gaps are uncorrelated with model size. Therefore, there is no evidence that larger models are more fair, i.e., rationale alignment does not become more equal for demographic groups. In the context of toxicity classification, work from \cite{https://doi.org/10.48550/arxiv.2108.01250} also hints that size is not well correlated with fairness of models.

\paragraph{Do distilled models align better?}
Knowledge distillation has been proven to be effective in model compression while maintaining model performance \cite{Gou_2021}. But can it also be effective in improving NLP fairness? \citet{https://doi.org/10.48550/arxiv.2201.08542} find a consistent pattern of toxicity and bias reduction after model distillation. \citet{chai2022fairness} show promising results when approaching fairness without demographics through knowledge distillation. \citet{10.1145/3278721.3278725} discuss the benefits of applying knowledge distillation to leverage model interpretability. Motivated by these findings, we take results from LRP to look closer into group-model rationale agreement for distilled models, which we show in Table~\ref{tab:distilled_tokenf1}. We find overall higher rationale agreement for distilled models. However, there is no evidence that distilled models are also more fair: Only {\tt minilm-l6-h384-uncased} has a smaller performance gap between the best and worst-off group for both metrics compared to the average.

\begin{table}[t!]
    \centering
    \resizebox{\columnwidth}{!}{
    \begin{tabular}{lcccc}
    \toprule
         ~ & $\mathrm{\text{token-}F_1}$ ($\uparrow$) &  $\mathrm{\text{IOU-}F_1}$ ($\uparrow$) & \makecell{min-max\\ $\mathrm{\text{token-}F_1}$ ($\downarrow$)} & \makecell{min-max\\ $\mathrm{\text{IOU-}F_1}$ ($\downarrow$)} \\ \midrule
         minilm-l6-h384-unc. & \bf{.31} & \bf{.28} & \bf{.045} &  \bf{.068}\\
         minilm-l12-h384-unc. & .27  & .21 & \bf{.045}& .083\\
         distilbert-base-unc. & \bf{.29}  & \bf{.24} & .064 & .100\\
         distilroberta-base & \bf{.36} & \bf{.36}  & .065& \bf{.069} \\ \midrule
         Avg. (16 models) & .29  & .24 & .054 & .081\\
         \bottomrule
    \end{tabular}}
    \caption{Group-model alignment for four distilled models. Bottom row shows average scores across all 16 models considered in this paper. Values in \textbf{bold} are better than the average (lower if $\downarrow$, higher if $\uparrow$). While rationale alignment ($\mathrm{\text{token-}F_1}$ and $\mathrm{\text{IOU-}F_1}$) seem to be better for distilled models, only {\tt minilm-l6-h384-uncased} is also fairer than the average (in terms of min-max difference) with both metrics.}
    \label{tab:distilled_tokenf1}
\end{table}

\section{Conclusion}
In this paper, we present a new collection of three existing datasets with demographics-augmented annotations, balanced across age and ethnicity. By having annotators choose the right label and marking supporting evidence for their choice, we find that what counts as a rationale differs depending on peoples' socio-demographic backgrounds. 

Through a series of experiments with 16 popular model architectures and two families of explainability methods, we show that model rationales align better with older individuals, especially on sentiment classification. We look closer at model size and the influence of distilled pretraining: despite the fact that larger models perform better in general NLP tasks, we find negative correlations between model size and rationale agreement. Furthermore, from the point of view of performance parity, we find no evidence that increasing model size improves fairness. Likewise, distilled models do not seem to be more fair in terms of rationale agreement, however they do present overall higher scores. 

This work indicates the presence of undesired biases that {\em do not necessarily surface in task performance}. 
We believe this provides an important addendum to the fairness literature: Even if models are fair in terms of predictive performance, they may still exhibit biases that can only be revealed by considering model rationales. If models are equally right, but only right for the right reasons in the eyes of some groups rather than others, they will likely be less robust for the latter groups.

\section*{Limitations}

Our analysis is limited to non-autoregressive 
Transformer-based models, fine-tuned with the same set of hyperparameters. Hyperparameter optimization would undoubtedly lead to better performance for some models, but we 
fine-tuned each model with standard hyperparameter values for solving sentiment analysis tasks \cite{deyoung2019eraser} to reduce resource consumption. This should not affect the conclusions drawn from our experiments. 

Comparing human rationales and rationales extracted with interpretability methods such as Attention Rollout and LRP is not straightforward. 
Overall agreement scores depend on how model rationales are converted into categorical values (top-$k^{gd}$). See \citet{joergensen2022} for discussion.

\section*{Acknowledgments}
Many thanks to Stephanie Brandl, David Dreyer Lassen, Frederik Hjort, Emily Pitler and David Jurgens for their insightful comments.

This work was supported by the Novo Nordisk Foundation.

\section*{Ethics Statement}

\paragraph{Broader impact}
Although explainability and fairness are broadly viewed as intertwined subjects, very little work has studied the two concepts together \citep{10.1145/3301275.3302265,gonzalez-etal-2021-explanations,ruder-etal-2022-square}. This study is a first of its kind to examine fairness issues of explainability methods and to publish human rationales with diverse socio-demographic information. We hope this work will impact the NLP research community towards more data-aware and multi-dimensional investigations of models and methods, and towards further studies of biases in NLP.

\paragraph{Personal and sensitive data} This study deals with personal and sensitive information. The responses are anonymous and cannot be used to identify any individual.

\paragraph{Informed consent} The participants were informed of the study's overall aim, the procedure and confidentiality of their responses. With this information, the participants consented to the use and sharing of their responses.

\paragraph{Potential risks} We do not anticipate any risks of participation in the study, yet we do note a recent awareness of poor working conditions among crowdworkers for AI data labeling in some countries \citep{AIethics}. The recruitment platform Prolific, used in this study, is targeted towards research (rather than AI development) and has stricter rules on participant screening and minimum wages \citep{Palan2017ProlificacASP}, compared to other popular platforms, which we hope reduce the risk of such poor working conditions.

\paragraph{Remuneration}
The participants were paid an average of $7.1$\textsterling/hour ($\approx8.8$\$/hour).

\paragraph{Intended use} The collected annotations and demographic information will be publicly available to be used for research purposes only.

\bibliography{anthology,custom}
\bibliographystyle{acl_natbib}

\clearpage
\appendix

\label{sec:appendix}

\section{Annotation guidelines and task examples}
\label{app:annotation-guidelines}

On the next pages, we firstly show the annotation instructions given to annotators within the Qualtrics surveys. Full exports of the surveys are available in our GitHub repository.\footnote{\url{https://github.com/terne/Being_Right_for_Whose_Right_Reasons}.} 

We created instructions specific for each dataset (DynaSent, SST-2, and CoS-E), leaning on prior work of annotating labels and rationales for these and similar datasets \cite{potts-etal-2020-DynaSent,zaidan-etal-2007-using,deyoung2019eraser}, as described in the paper, section \ref{sec:annotation_process}. 

\begin{figure*}[b!]
\begin{framed}
\small

\textbf{Instructions}

\noindent Please read these instructions carefully.\\
 
\noindent You will be shown 10 sentences from reviews of products and services. For each, your task is to choose from one of our three labels:\\
 
\noindent Positive: The sentence conveys information about the author's positive evaluative sentiment.\\ 	

\noindent Negative: The sentence conveys information about the author's negative evaluative sentiment.\\ 	

\noindent No sentiment: The sentence does not convey anything about the author's positive or negative sentiment.\\  

\noindent Here are some examples of the labels:\\
 
\noindent Sentence: This is an under-appreciated little gem of a movie.\\
\noindent (This is Positive because it expresses a positive overall opinion.)\\
 
\noindent Sentence: I asked for my steak medium-rare, and they delivered it perfectly!\\
\noindent (This is Positive because it puts a positive spin on an aspect of the author's experience.)\\
 
\noindent Sentence: The screen on this device is a little too bright.\\
\noindent (This is Negative because it negatively evaluates an aspect of the product.)\\
 
\noindent Sentence: The book is 972 pages long.\\
\noindent (This is No sentiment because it describes a factual matter with not evaluative component.)\\
 
\noindent Sentence: The entrees are delicious, but the service is so bad that it's not worth going.\\
\noindent (This is Negative because the negative statement outweighs the positive one.)\\
 
\noindent Sentence: The acting is great! The soundtrack is run-of-the mill, but the action more than makes up for it.\\
 (This is Positive because the positive statements outweighs the negative.)\\

\noindent We further ask you to specify what snippets of text, in the sentence, you think acts as supporting evidence for your chosen label. The sentence will be shown to you as illustrated below, and your task is to mark (by clicking on them) all the words you think shows evidence for the sentiment label you chose.

\includegraphics[width=0.7\linewidth]{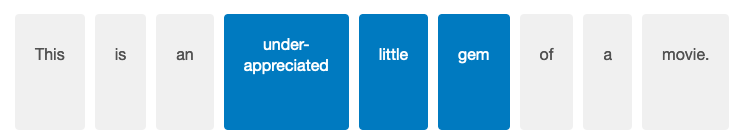}

\noindent Be aware that some sentences might be too long to fit on your screen. You therefore have to remember to scroll in order to see all the words that can be marked as evidence.\\  

\noindent Click the forward button below when you are ready to start the task.

\end{framed}
\caption{DynaSent annotation instructions.}
\label{fig:dynasentinstructions}
\end{figure*}

\begin{figure*}[b!]
\begin{framed}
\small
\textbf{Instructions}

\noindent Please read these instructions carefully.\\
 
\noindent You will be shown approximately 10 sentences from reviews of movies. For each, your task is to choose from one of our three labels:\\ 

\noindent Positive: The sentence conveys information about the author's positive evaluative sentiment.\\      	\noindent Negative: The sentence conveys information about the author's negative evaluative sentiment.\\      	
\noindent No sentiment: The sentence does not convey anything about the author's positive or negative sentiment.\\  
 
\noindent Here are some examples of the labels:\\

\noindent Sentence: This is an under-appreciated little gem of a movie.\\
\noindent (This is Positive because it expresses a positive overall opinion.)\\
 
\noindent Sentence: he is one of the most exciting martial artists on the big screen, continuing to perform his own stunts and dazzling audiences with his flashy kicks and punches.\\
\noindent (This is Positive because it positively evaluates an aspect of the movie.)\\
 
\noindent Sentence: The acting is great! The soundtrack is run-of-the-mill, but the action more than makes up for it.\\
\noindent (This is Positive because the positive statements outweigh the negative.)\\

\noindent Sentence: The story is interesting but the movie is so badly put together that even the most casual viewer may notice the miserable pacing and stray plot threads.\\
\noindent  (This is Negative because the negative statement outweighs the positive one.)\\

\noindent Sentence: A woman in peril. A confrontation. An explosion. The end. Yawn. Yawn. Yawn.\\
\noindent (This is Negative because it puts a negative spin on the author's experience.)\\
 
\noindent Sentence: don’t go see this movie.\\
\noindent (This is Negative because it recommends against seeing the movie, reflecting a negative evaluation.)\\

\noindent Sentence: it is directed by Steven Spielberg.\\
\noindent (This is No sentiment because it describes a factual matter with no evaluative component.)\\
 
\noindent Sentence: I saw it in the local theater with my best friend.\\
\noindent (This is No sentiment because it does not say anything about the movie.)\\

\noindent We further ask you to specify what snippets of text, in the sentence, you think acts as supporting evidence for your chosen label. The sentence will be shown to you as illustrated below, and your task is to mark (by clicking on them) all the words you think shows evidence for the sentiment label you chose.
 
\includegraphics[width=0.7\linewidth]{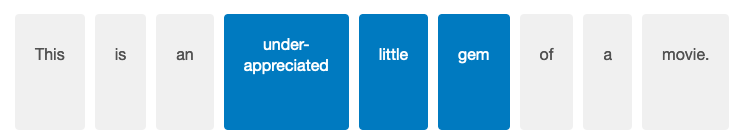}

\noindent Be aware that some sentences might be too long to fit on your screen. In that case you have to scroll in order to see all the words that can be marked as evidence.\\  

\noindent Click the forward button below when you are ready to start the task.

\end{framed}
\caption{SST-2 annotation instructions.}
\label{fig:sstinstructions}
\end{figure*}

\begin{figure*}[b!]
\begin{framed}
\small

\textbf{Instructions}\\
\noindent (Please read these instructions carefully.)\\
   
\noindent You will be shown 10 multiple-choice questions. All questions and their answer-options have been created by other crowdworkers, who where instructed to create questions that can be fairly easily answered by humans without context, by the use of common-sense knowledge.\\

\noindent Your task is to firstly select the answer you think is most correct and sensible. We call this the label of the question. Secondly, we ask you to mark relevant words in the question that justifies your choice. Specifically, for each word in the question, if you think that removing it will decrease your confidence toward your chosen label, you should mark it.\\ 
   
\noindent In the image below, you see an example of how the task will be presented to you. To the question "Where do you find the most amount of leafs?", the option "Forest" is selected as the correct answer and four words have been marked as justification.

\includegraphics[width=0.7\linewidth]{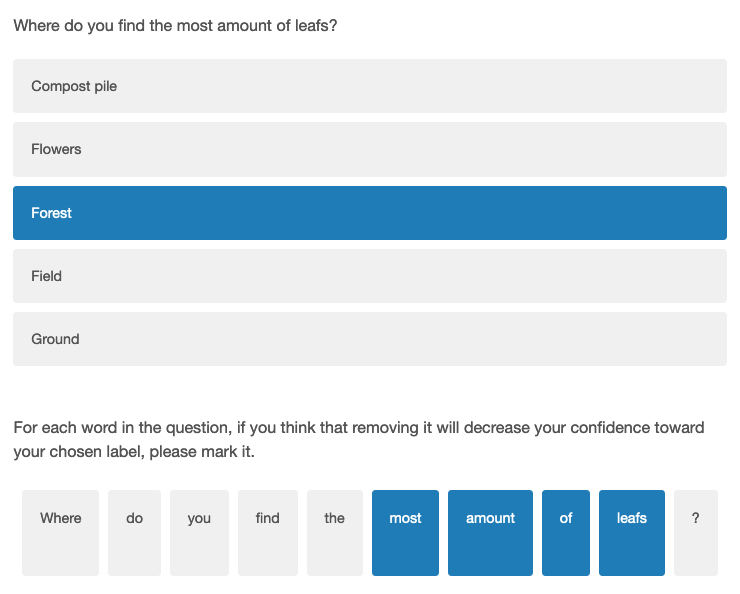}

\noindent When marking words, be aware that some questions might be longer and not fit perfectly on your screen. In that case you have to scroll in order to see all the words that can be marked. Also, the texts may have misspellings, typos and wrongly put spaces before punctuation – pay no attention to this.\\

\noindent Click the forward button below when you are ready to start the task.

\end{framed}

\caption{CoS-E annotation instructions.}
\label{fig:coseinstructions}
\end{figure*}

\begin{figure*}
    \begin{subfigure}[b]{\linewidth}
        \centering
    \includegraphics[width=0.7\linewidth]{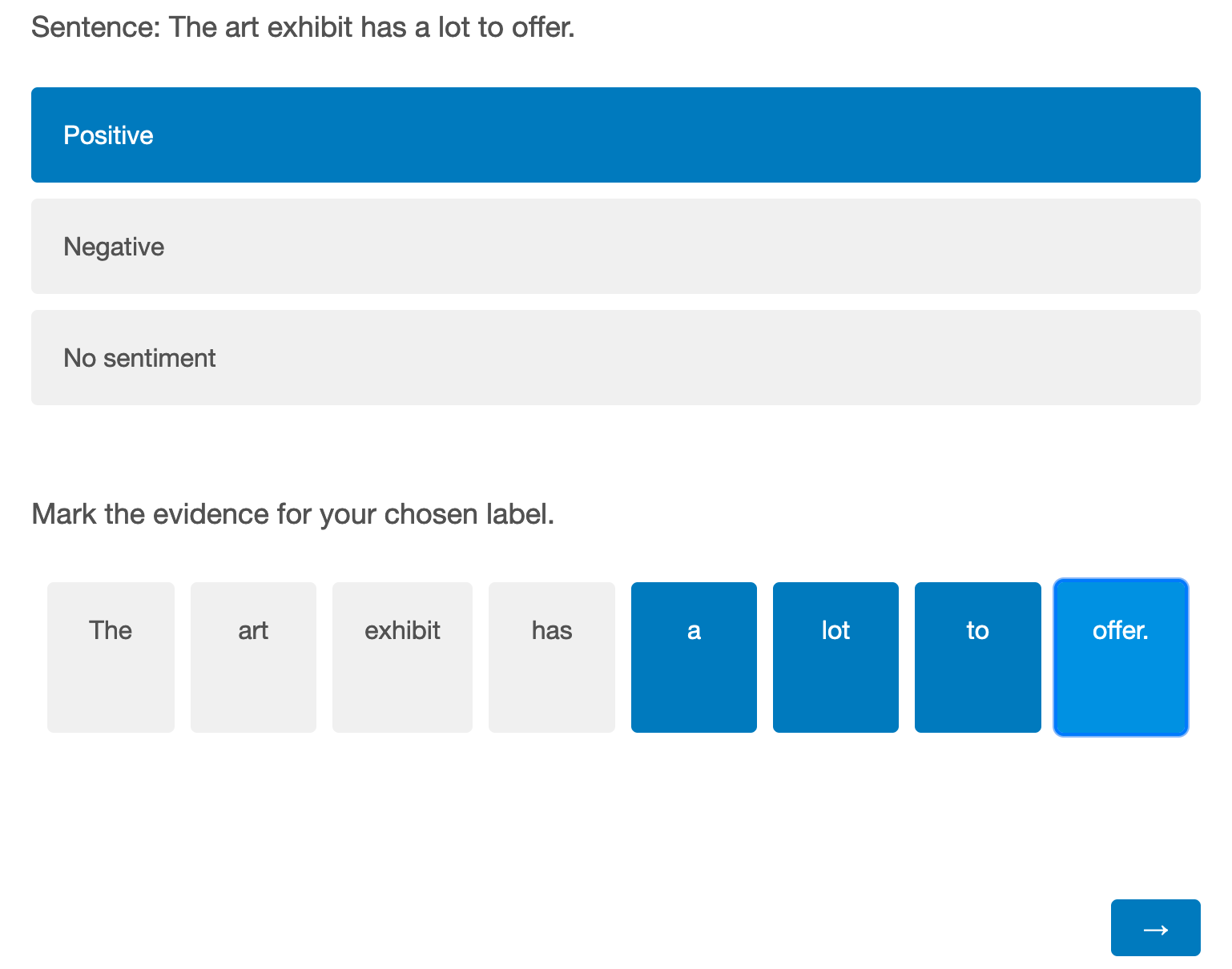}
    \caption{Sentiment annotation example.}
    \label{fig:my_label1}
    \end{subfigure}
    
    \begin{subfigure}[b]{\linewidth}
        \centering
    \includegraphics[width=0.7\linewidth]{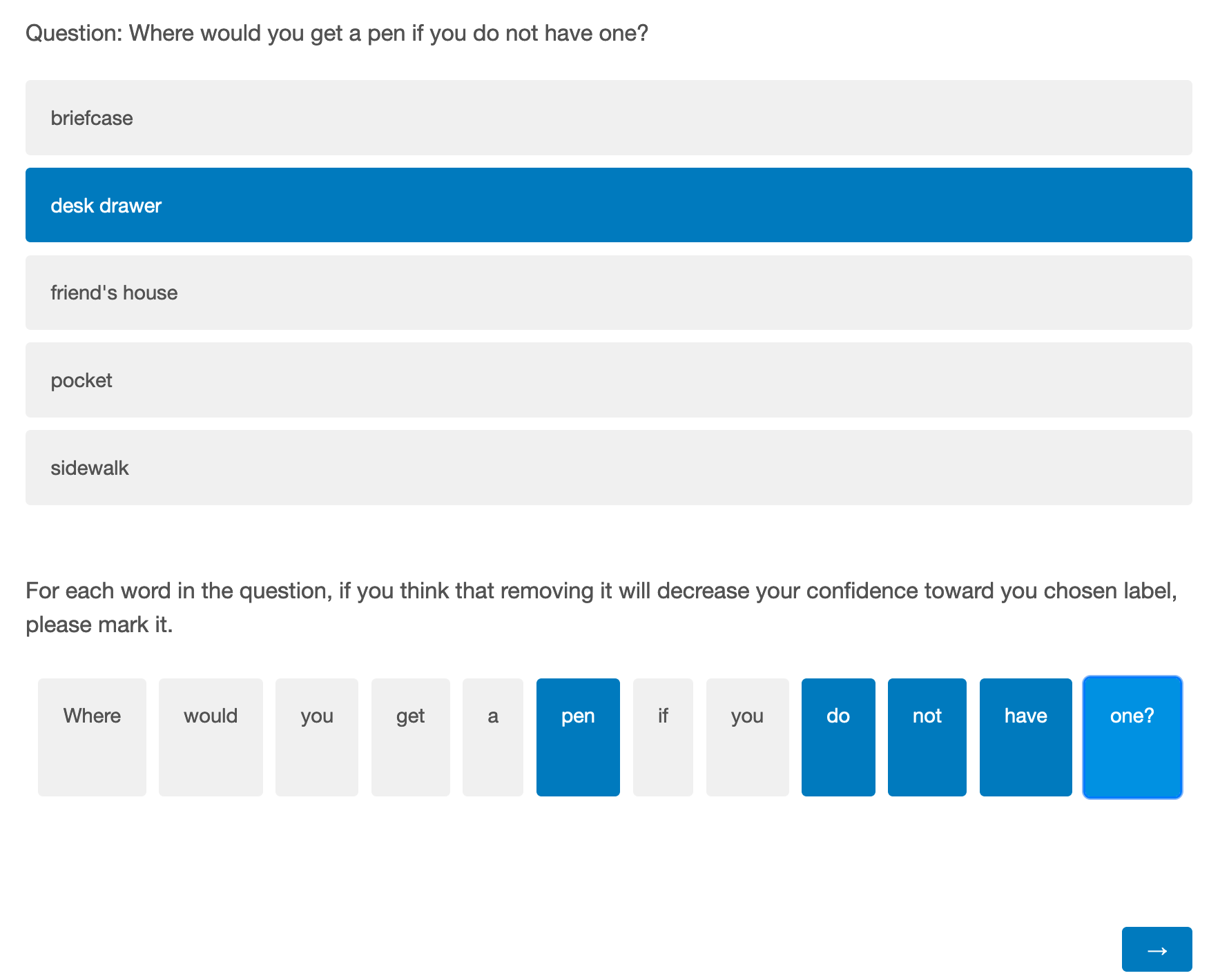}
    \caption{Common-sense reasoning annotation example.}
    \label{fig:my_label2}
    \end{subfigure}
    \caption{Screenshots of the annotation tasks as they are viewed in Qualtrics surveys.}
    \label{fig:twoexamples}
\end{figure*}

Figure \ref{fig:dynasentinstructions}, \ref{fig:sstinstructions}, and \ref{fig:coseinstructions} shows the instructions for DynaSent, SST-2 and CoS-E, respectively, and Figure \ref{fig:twoexamples} shows an example of how an instance for the sentiment task and the common-sense reasoning task is annotated, i.e. how it looked from the perspective of the crowdworkers.

Annotating rationales for the common-sense reasoning task is somewhat more complex than annotating rationales for sentiment: while we can ask annotators to mark `evidence' for a sentiment label -- often resulting in marking words that are positively or negatively loaded -- we cannot as simply ask for `evidence' for a common-sense reasoning answer without risking some confusion. Take, for instance, the question ``Where do you find the most amount of leafs?'' with the answer being `Forest', as shown in Figure \ref{fig:coseinstructions}. Here, the term 'evidence' might be misunderstood as actual evidence for why there would be more leafs in the forest compared to a field -- evidence which cannot be found within the question itself. We therefore re-phrase the rationale annotation instructions for CoS-E, following an example from \citet{chenghan2022}, and ask, ``For each word in the question, if you
think that removing it will decrease your confidence toward your chosen label,
please mark it.'' Furthermore, the subset of the CoS-E dataset, that we re-annotate, consists of the more `difficult' split of the CommonsenseQA dataset \cite{talmor-etal-2019-commonsenseqa, deyoung2019eraser}. To make the task as clear as possible to the annotators, we explain, in the instructions, that the question and answer-options have been created by other crowdworkers who were instructed to create questions that could be ``easily answered by humans without context, by the use of common-sense knowledge'', as is described by \citet{talmor-etal-2019-commonsenseqa}.




\begin{table}[h]
\small
    \centering
    \begin{tabular}{l|c|cccc}
    \toprule
        & &  \multicolumn{4}{c}{{\sc Complete label agreement}} \\
         {\sc Dataset} & {\sc N} & {\sc Pos} & {\sc Neg} & {\sc Neutral} &  {\sc Total}\\
         \midrule
         DynaSent & 480 & 105 & 102 & 2& 209\\
         SST & 263 & 79 & 73 & 0 & 152\\
         CoS-E  &  500 & - & - & - & 161 \\
    \bottomrule
    \end{tabular}
    \caption{Number of instances, in our (re-)annotated data, where all annotator groups agreed upon the instance's label.}
    \label{tab:commonlabs}
\end{table}

\section{Annotations Overview}

Table \ref{tab:annotators} gives further information on the distribution of annotators, across groups and datasets, as well as ratios of rationale lengths to input lengths. Table \ref{tab:commonlabs} shows the number of instances in the data subsets, we work with, and the number of instances where all our annotator groups agreed on the label and that are therefore used for rationale-agreement analyses.

\label{app:annotations-overview}

\begin{table*}[!htbp]
\footnotesize
    \centering
    \begin{tabular}{llcccccccc}
        \toprule
         {\sc Dataset}&&{\sc BO}&{\sc BY}&{\sc LO}&{\sc LY}&{\sc WO}&{\sc WY}& {\sc Total/Avg.}\\
         \midrule
         \multirow{3}{*}{DynaSent}
         & Annot. &51&56&61&73&54&51&346\\
         & Annot.$^{*}$ &48 (58\%F) &48 (67\%F) &48 (44\%F) &48 (40\%F)&48 (56\%F)&48 (48\%F)&288\\
         & RLR & 33.7& 32.5& 31.5& 29.8& 34.7& 29.1 & 31.9\\
         \midrule
         \multirow{3}{*}{SST} 
         & Annot. &28&27&53&43&27&29&207\\
         & Annot.$^{*}$ &26 (69\%F)&26 (58\%F)&26 (38\%F)&26 (31\%F)&26 (38\%F)&26 (69\%F) &156\\
         & RLR  & 32.1& 25.1& 30.7& 27.8& 29.1& 32.7 & 29.6 \\
         \midrule
         \multirow{3}{*}{CoS-E}
         & Annot. &52&56&74&85&54&55&376\\
         & Annot.$^{*}$ &50 (60\%F)&50 (60\%F)&50 (40\%F)&50 (48\%F)&50 (48\%F)&50 (40\%F) &300\\
         & RLR &31.9& 32.9& 34.1& 32.2& 33.3& 33.6 & 33.0\\
        \bottomrule 
    \end{tabular}
    \caption{Overview of our annotated data. Rows display statistics per dataset. Columns refer to each demographic group: Black/African American old (BO) and young (BY), Latino/Hispanic old (LO) and young (LY), White/Caucasian old (WO) and young (WY). Last column show the total quantity of each feature over all groups. Row-wise within each dataset: `Annot.' and `N' reflect the total number of annotators and instances, respectively. Annot.$^{*}$ refers to the number of annotators left after pre-processing (see exclusion criteria in Section \ref{sec:recruitment}). 
    Number shown between brackets refers to the percentage of female annotators. RLR represents the ratio of rationale length to its input length (percentage).
    }
    \label{tab:annotators}
\end{table*}


\section{Supplementary Figures}
\label{app:supplementary-figures}

For completeness, we provide supplementary figures for all the metrics and datasets analyzed in the paper. 

\subsection{Label Agreement}
\begin{figure*}[ht!]
    \centering
    \includegraphics[width=0.6\textwidth]{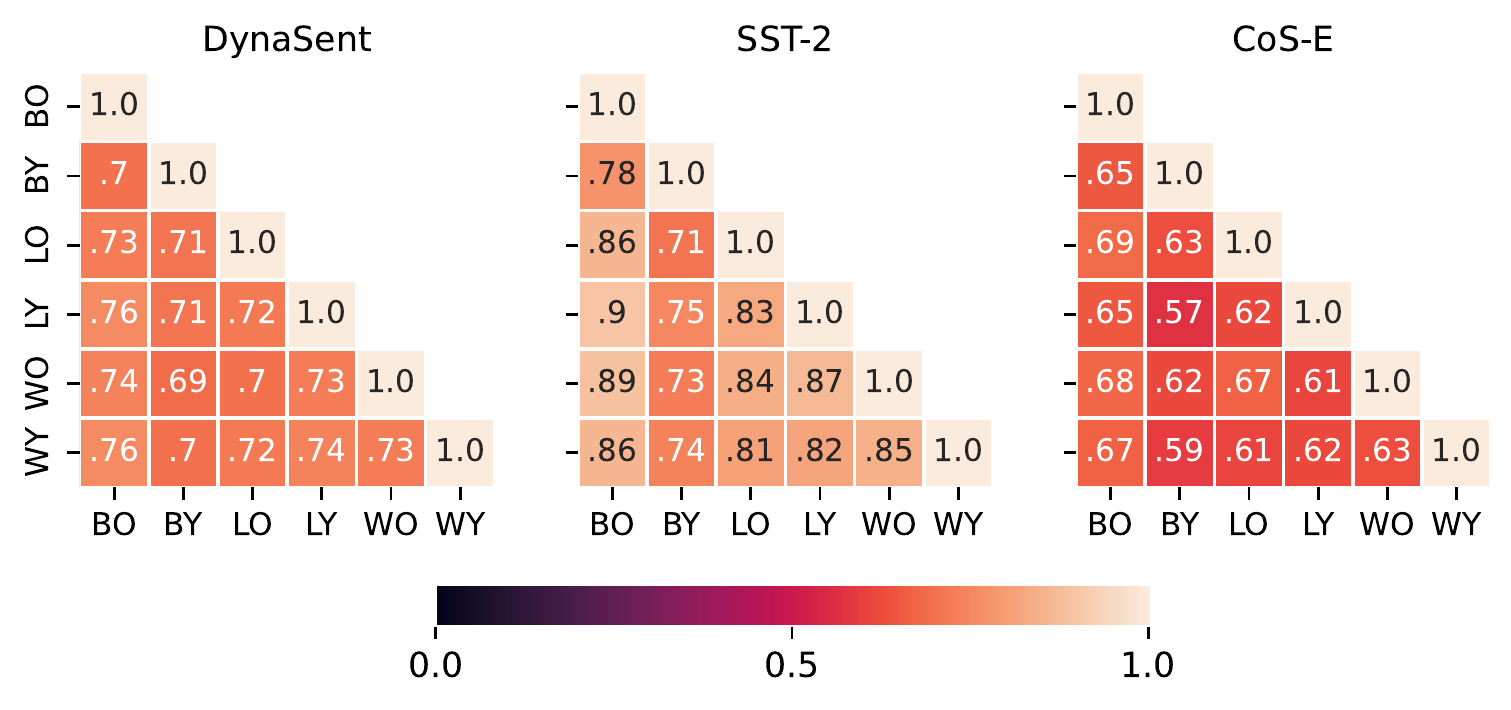}
    \caption{Group-group label agreement (F1-scores).}
    \label{fig:group_group_f1}
\end{figure*}

Heatmaps in Figure~\ref{fig:group_group_f1} show the level of group-group label agreement across datasets. Similar to what is shown in Figure~\ref{fig:groupgroup_agreement}, BY consistently exhibit lower level of agreement.

\begin{figure*}[ht!]
    \centering
    \includegraphics[width=0.8\textwidth]{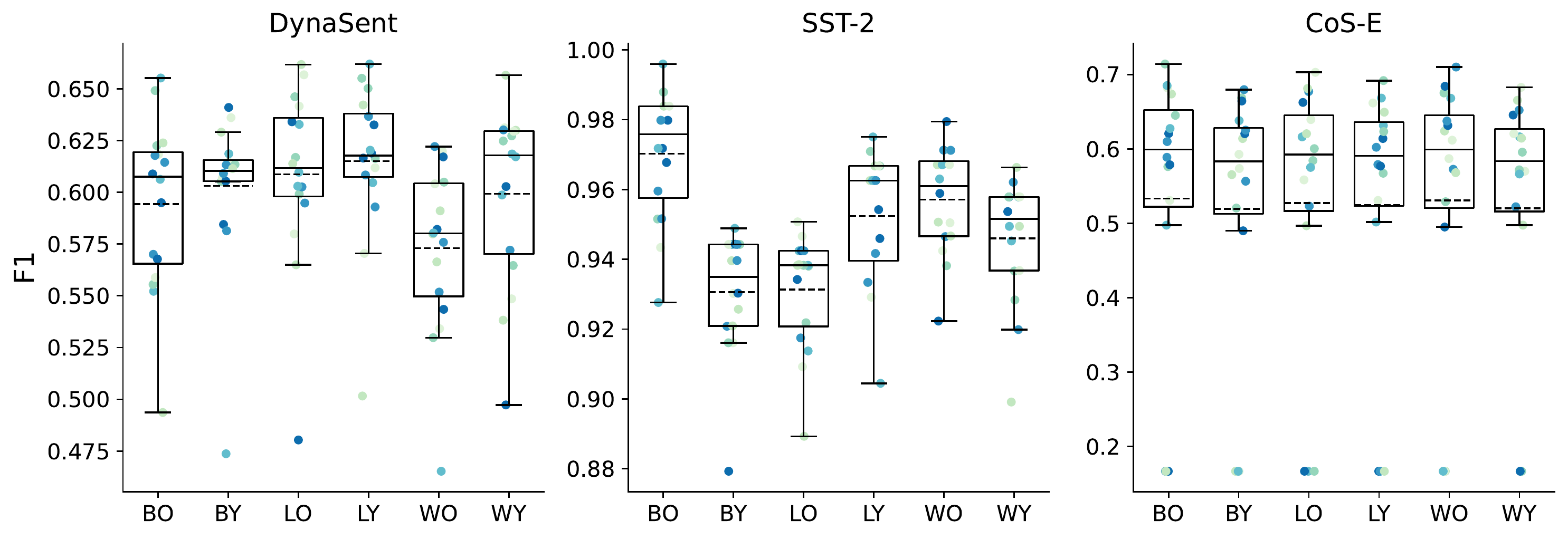}
    \caption{Group-model label agreement (F1-scores).}
    \label{fig:model_group_f1}
\end{figure*}

Box-plots in Figure~\ref{fig:model_group_f1} represent group-model label agreement. Each dot represents the F1-score of each model. While for Cos-E the models generally exhibit lower variability across groups, the level of agreement is also lower (as shown in Figure~\ref{fig:f1-scsores}).

\subsection{Rationale Alignment}
\label{app:rationale-agreement}
\begin{figure*}
    \centering
    \begin{subfigure}[b]{0.46\textwidth}
         \centering
         \includegraphics[width=\textwidth]{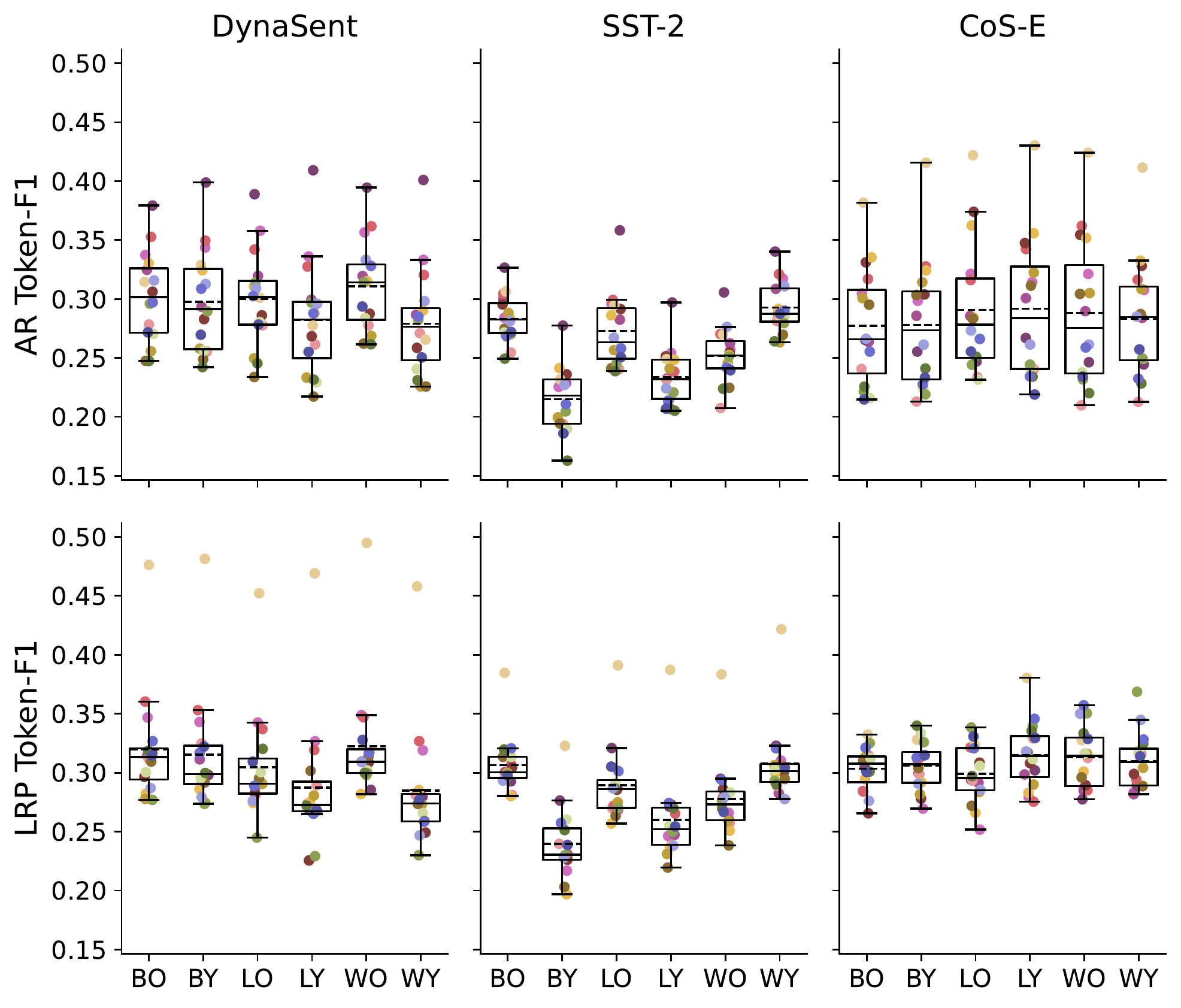}
         \caption{Token-F1 scores}
     \end{subfigure}
     \begin{subfigure}[b]{0.53\textwidth}
         \centering
         \includegraphics[width=\textwidth]{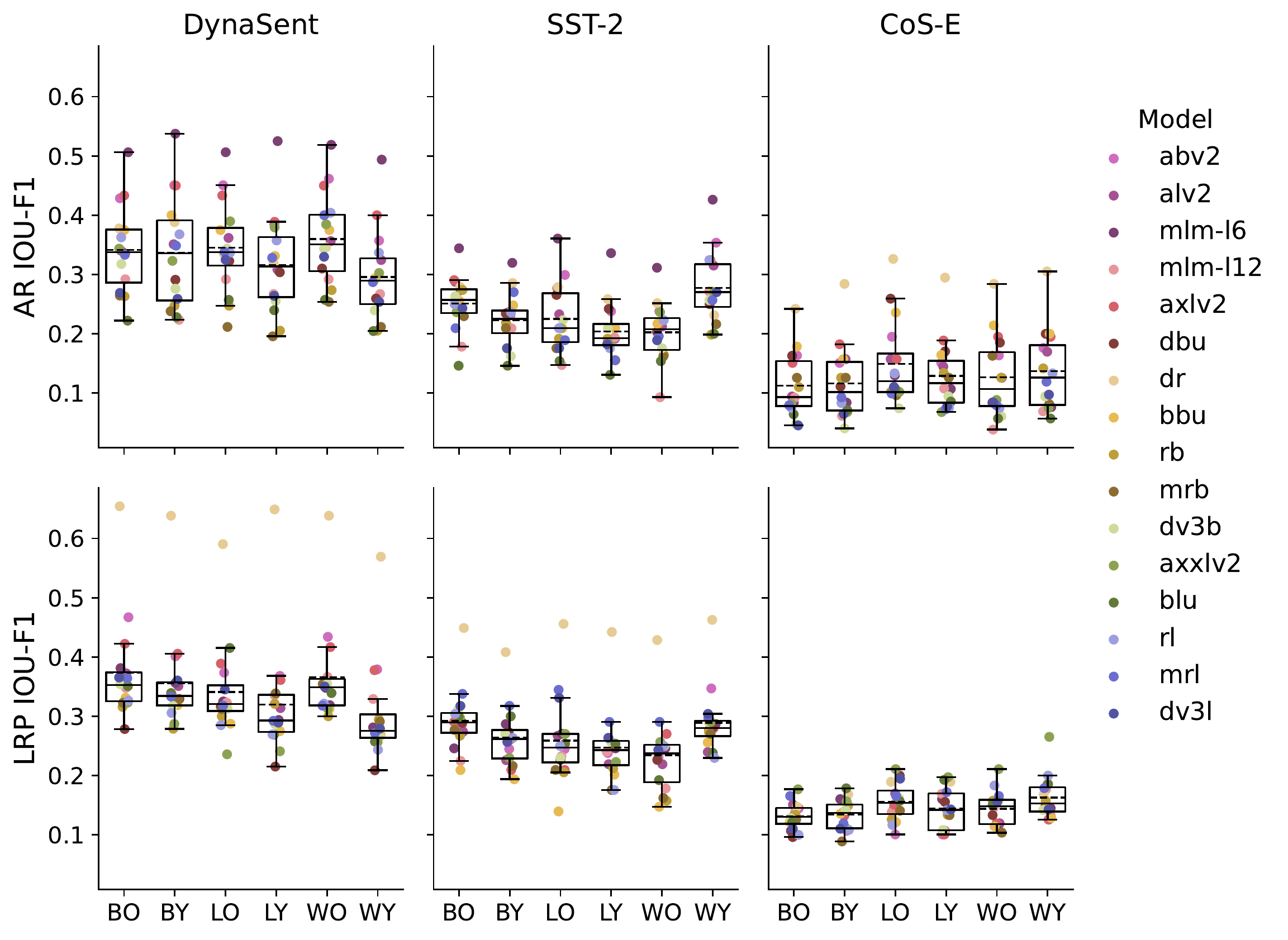}
         \caption{IOU-F1 scores}
     \end{subfigure}
    \caption{Box-plots of group-model rationale agreement for the each dataset measured with Token-F1 (left) and IOU-F1 (right). Model rationales are extracted with Attention Rollout (top row) and LRP (bottom row). Each dot represents a model's agreement with the respective group.}
    \label{fig:complete-group-model-alignment}
\end{figure*}

Figure \ref{fig:complete-group-model-alignment} is the extended version of Figure \ref{fig:tokenf1_boxplot}, showing the group-model rationale agreement for each dataset, each explainability method and with two metrics for measuring agreement, $\mathrm{\text{token-}F_1}$ and $\mathrm{\text{IOU-}F_1}$.

\begin{figure*}
    \centering
    \includegraphics[width=\textwidth]{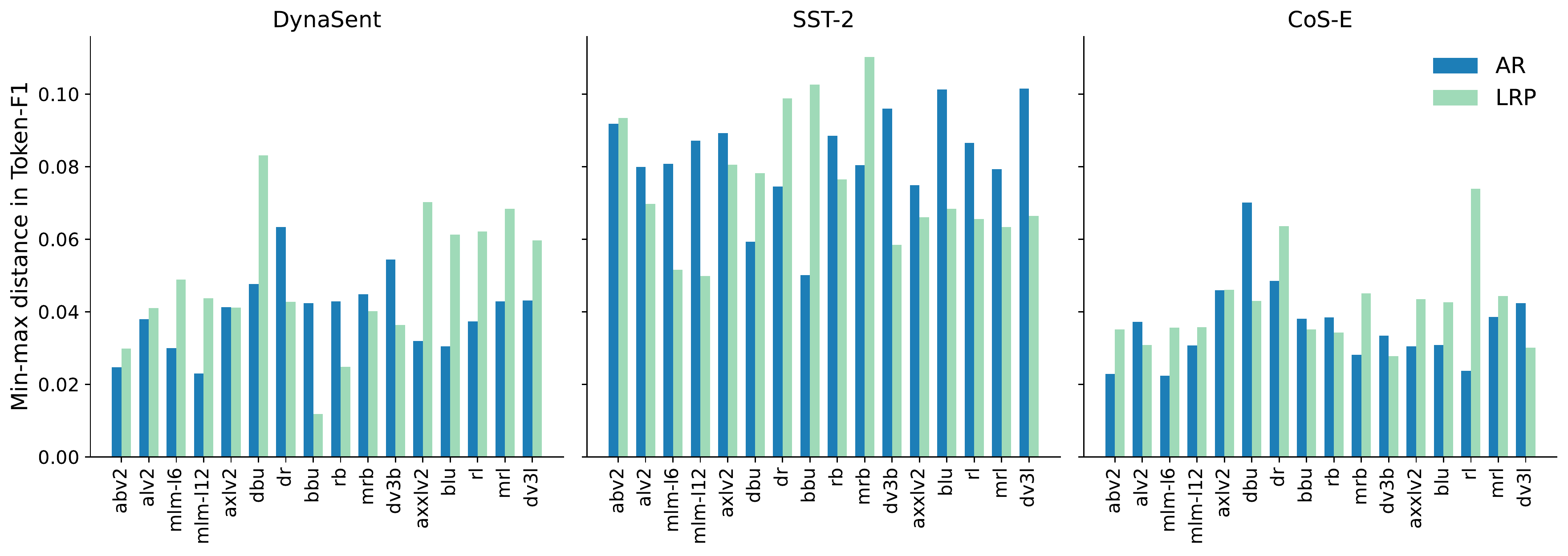}
    \caption{Per-model difference between the group with the lowest (min) and highest (max) model-group agreement measured with token-F1. Models on the x-axis are sorted by model size. The min-max captures a measure of fairness, with a smaller difference entailing more equal model-group rationale alignments. We find that the differences are uncorrelated with model size (in Million parameters), as is visible in this plot.}
    \label{fig:minmax}
\end{figure*}

The bar charts in Figure \ref{fig:minmax} shows, per model and dataset, the distance between the group with the lowest and highest agreement with the model (by $\mathrm{\text{token-}F_1}$), which we refer to as the ``min-max $\mathrm{\text{token-}F_1}$ gaps'' in section \ref{sec:analysis-modelgroup}. We include this plot because it serves to better illustrate the gaps themselves, and how they are uncorrelated with model size, compared to what Figure \ref{fig:modelsize_tokenf1} in the paper can convey.

\end{document}